\documentclass{article}

\PassOptionsToPackage{numbers,compress}{natbib}

\usepackage[preprint]{neurips_2026}

\usepackage[utf8]{inputenc} % allow utf-8 input
\usepackage[T1]{fontenc}    % use 8-bit T1 fonts
\usepackage{hyperref}       % hyperlinks
\usepackage{url}            % simple URL typesetting
\usepackage{booktabs}       % professional-quality tables
\usepackage{amsfonts}       % blackboard math symbols
\usepackage{nicefrac}       % compact symbols for 1/2, etc.
\usepackage{microtype}      % microtypography
\usepackage{xcolor}         % colors
\usepackage{graphicx}       % figures and resizebox
\usepackage{subcaption}     % subfigures/subcaptions
\usepackage{amsmath}        % math environments
\usepackage{float}
\usepackage{placeins}
\usepackage{amssymb}
\usepackage{multirow}
\usepackage{longtable}
\usepackage{array}
\usepackage{wrapfig}

\hypersetup{
  pdftitle={Not All Tokens Matter Equally: Dynamic In-context Vector Distillation with Decisive-Token Supervision for Long-form Medical Report Generation},
  pdfauthor={Ning Wu, Rui Liu, Xinkun Lin, Weixing Chen, Jinxi Xiang, Tao Wei, Lina Yao, Mingjie Li}
}

\title{Not All Tokens Matter Equally: Dynamic In-context Vector Distillation with Decisive-Token Supervision for Long-form Medical Report Generation}
\author{%
\begin{tabular}{c}
Ning Wu$^{1}$, Rui Liu$^{2}$, Xinkun Lin$^{1}$, Weixing Chen$^{3}$ \\
Jinxi Xiang$^{4}$, Tao Wei$^{5}$, Lina Yao$^{1,*}$, Mingjie Li$^{5,*}$ \\[1mm]
$^{1}$UNSW Sydney, Sydney, Australia \\
$^{2}$University of Technology Sydney, Sydney, Australia \\
$^{3}$School of Computer Science and Engineering, Sun Yat-sen University, China \\
$^{4}$Stanford University, Palo Alto, United States \\
$^{5}$Shanghai Jiao Tong University, Shanghai, China \\[1mm]
\texttt{wuning84@outlook.com, rliu0016@gmail.com, xinkun.lin124@gmail.com} \\
\texttt{chenwx228@mail2.sysu.edu.cn, xiangjx@stanford.edu, taowei@sjtu.edu.cn} \\
\texttt{lina.yao@unsw.edu.au, mingjie.li@sjtu.edu.cn} \\[1mm]
$^{*}$Corresponding authors.
\end{tabular}
}

\begin{document}

\maketitle

\begin{abstract}
Distilling demonstration effects into hidden-space interventions offers a lightweight alternative to full finetuning. However, existing multimodal variants are mostly evaluated on short-form tasks, where outputs end after a few tokens. Extending these methods to long-form generation exposes a fundamental yet underexamined limitation: token-level distillation implicitly treats all output tokens as equally informative, but long-form outputs are dominated by high-frequency template and grammatical tokens, while the tokens that actually determine output quality are sparsely distributed. In medical report generation (MRG), two such decisive tokens stand out: pathology-related tokens that determine diagnostic content, and the end-of-sequence (EOS) event that determines termination. Both receive insufficient supervision under uniform cross-entropy, and autoregressive decoding further compounds the problem by drifting away from teacher-forced trajectories. We propose DIVE, a frozen-backbone distillation framework that addresses long-form report generation through two complementary mechanisms matched to these failures. Decisive-token supervision restores supervision balance by upweighting the cross-entropy contribution of pathology-related tokens and the EOS event, ensuring that content fidelity and termination are learned during training rather than imposed at decoding time. State-conditioned dynamic steering replaces fixed open-loop residuals with hidden-state-dependent adapters, allowing the injected signal to adapt as decoding drifts. Experiments on MIMIC-CXR and CheXpert Plus with two medical VLM backbones show that DIVE consistently ranks among the strongest methods across lexical and clinical-proxy metrics. Our method achieves the best BLEU-4, ROUGE-L, and RadGraph F1 in all dataset--backbone settings, while remaining competitive on coarse label-level CheXbert F1.
\end{abstract}

\section{Introduction}

In-context learning (ICL) enables large models to adapt to new tasks through demonstrations without parameter updates
\citep{brown2020language,min2022rethinking,xie2022bayesianicl,garg2022what,akyurek2023what,vonoswald2023transformers,dong2024surveyicl}.
Recent work further suggests that demonstration-induced behavior can be compressed into latent or parameter-space interventions, including in-context vectors, task arithmetic, activation steering, representation engineering, and multimodal task vectors
\citep{liu2023icv,ilharco2023taskarithmetic,turner2023activationaddition,zou2023representationengineering,li2023iti,rimsky2024activationsteering,peng2024live,huang2024mtv}.
These methods reduce prompt overhead and provide lightweight alternatives to full finetuning, but most multimodal evidence remains concentrated on short-form tasks such as visual question answering, where outputs end after only a few tokens.
This raises a basic question for the long-form regime: \emph{can a distilled hidden-space intervention remain reliable when generation must be sustained over many tokens and when output quality depends on getting a few decisive tokens right rather than on matching every token equally well?}
% the model must decide not only what to say, but also \emph{when to stop}?

We study this question in medical report generation (MRG), a long-form multimodal task where a model generates a multi-sentence radiology report from given medical images~\citep{wang2025cxpmrg,park2025dart,liu2025mlrg,heiman2025factchexcker,wang2025curv,cheng2026rethinking}. We treat MRG not as a clinical-deployment target, but as a representative long-form multimodal stress test: outputs are long enough for autoregressive errors to accumulate, and, crucially, their quality is highly non-uniform across token positions. A radiology report is dominated by template phrases (``the lungs are clear'', ``there is no evidence of'') that are easy to predict from local context, while the tokens that actually determine output quality are sparsely distributed. Two such decisive tokens stand out: pathology-related tokens (e.g., ``pneumothorax'', ``consolidation'', ``cardiomegaly'') that determine the diagnostic content of the report, and the end-of-sequence (EOS) event that determines whether the report terminates at the appropriate boundary. Late termination is particularly costly in this domain, as continuing past the visually supported content can introduce statements that are not grounded in the image.

This non-uniform structure exposes a fundamental limitation of token-level distillation: standard cross-entropy weights all output tokens equally, but the value of each token to overall report quality is far from equal. As a result, the gradient signal is dominated by abundant template tokens that are already easy to predict, while the rare but decisive tokens (pathology mentions and the EOS event) receive insufficient supervision. We refer to this phenomenon as decisive-token under-supervision. This issue is related to long-standing issues in sequence generation, including train--test mismatch under teacher forcing, length bias, beam-search degradation, and degeneration in autoregressive decoding~\citep{bengio2015scheduled,ranzato2016sequence,wiseman2016beamsearch,murray2018length,cohen2019beam,meister2020beam,holtzman2020curious,welleck2020unlikelihood}, but is specifically sharpened in the token-level distillation setting, where supervision density is uniform across positions by construction. At inference time, autoregressive decoding further compounds the problem by moving the student away from teacher-forced hidden-state trajectories, making fixed open-loop steering brittle over long horizons.

\begin{figure}[t]
    \centering
    \includegraphics[width=1.0\linewidth]{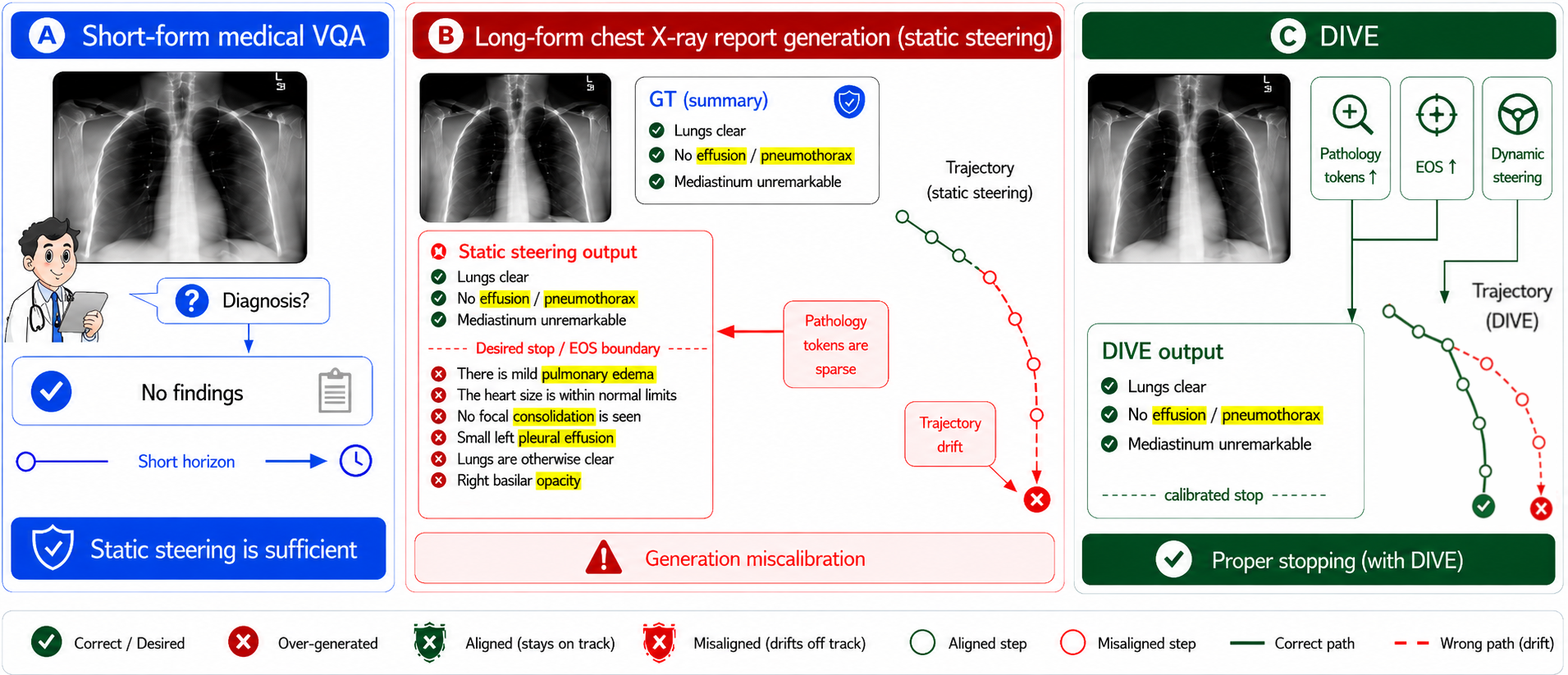}
\caption{
\textbf{Why short-form distilled steering does not directly transfer to long-form medical report generation.}
Static steering is effective for short-form medical VQA because outputs are short and termination is largely template-driven (A).
In long-form chest X-ray report generation, sparse pathology-related tokens and the EOS boundary are easily overwhelmed by frequent template tokens, and autoregressive drift further leads to miscalibrated continuation (B).
DIVE mitigates these failures through decisive-token supervision and state-conditioned dynamic steering, improving stopping calibration and clinical faithfulness (C).
}
    \label{fig:stopping_failure}
\end{figure}

We propose \textbf{DIVE}, a frozen-backbone distillation framework for long-form MRG that targets these two failure sources with complementary mechanisms.
% DIVE distills a demonstration-augmented teacher into query-only dynamic steering modules, while keeping the base vision-language model frozen.
% The method combines two mechanisms matched to the diagnosed failures.
Its primary mechanism, \emph{decisive-token supervision,} explicitly upweights the cross-entropy contribution of the two decisive token classes during distillation: pathology-related tokens, identified through a domain-prior vocabulary covering CheXpert disease categories and their common clinical phrasings, and the EOS event. This raises the effective training signal for tokens that actually determine report quality, addressing supervision density imbalance during training rather than imposing it at decoding time. 
% includes the EOS token in the supervised answer span and upweights its contribution to the distillation objective. This raises the effective training signal for the rare termination event, ensuring that stopping is learned during training rather than imposed at decoding time.
Complementing this, \emph{state-conditioned dynamic steering} replaces fixed open-loop residuals with a hidden-state-dependent intervention, so that the injected signal is computed from the state actually visited at each decoding step rather than from a single offline-estimated direction. 
The two mechanisms target distinct sources of failure, supervision-density imbalance and trajectory drift, and we find empirically that they contribute to different aspects of the final behavior.
% Together, these mechanisms turn static latent steering into a closed-loop form of distilled adaptation: the intervention depends on where the model currently is in generation, while the objective explicitly teaches when generation should end.

We evaluate DIVE on MIMIC-CXR~\citep{johnson2019mimic} and CheXpert Plus~\citep{chambon2024chexpertplus} using both lexical and clinical efficacy  metrics.
Within this scope, DIVE provides a stronger overall trade-off than zero-shot prompting, vanilla ICL~\citep{brown2020language}, QLoRA finetuning~\citep{dettmers2023qlora}, and static ICV-style steering~\citep{liu2023icv,peng2024live}, improving long-form report quality while reducing miscalibrated over-generation.
Ablations further show a clear separation of roles between the two mechanisms: decisive-token supervision primarily improves clinically salient content and termination behavior, whereas dynamic steering contributes complementary gains in report-level generation quality.

Our contributions are threefold:
\begin{itemize}
    \item We identify decisive-token under-supervision as a key bottleneck when token-level distillation is extended from short-form tasks to long-form MRG, and show that diagnostic fidelity and termination calibration are two instances of the same underlying failure.
    \item We propose DIVE, a frozen-backbone distillation framework that combines decisive-token supervision with state-conditioned dynamic steering to address supervision-density imbalance and autoregressive trajectory drift.
    \item On MIMIC-CXR and CheXpert Plus, DIVE consistently improves over static ICV-style steering and achieves the strongest BLEU-4, ROUGE-L, and RadGraph F1 across all dataset--backbone settings, while maintaining competitive CheXbert performance.
\end{itemize}

\section{Related Work}

\paragraph{Distilling demonstration effects into lightweight interventions.}
In-context learning adapts a model through input demonstrations, but retaining demonstrations at inference increases context length and computational cost
\citep{brown2020language,min2022rethinking,xie2022bayesianicl,garg2022what,akyurek2023what,vonoswald2023transformers,dong2024surveyicl}.
This has motivated methods that compress task- or demonstration-induced behavior into lightweight interventions, including in-context vectors, task arithmetic, activation steering, inference-time intervention, representation engineering, and multimodal task vectors
\citep{liu2023icv,ilharco2023taskarithmetic,turner2023activationaddition,li2023iti,zou2023representationengineering,rimsky2024activationsteering,peng2024live,huang2024mtv}.
These approaches are related to parameter-efficient adaptation methods such as adapters, prefix tuning, prompt tuning, LoRA, and QLoRA
\citep{houlsby2019adapters,li2021prefix,lester2021prompttuning,liu2022ptuningv2,hu2022lora,dettmers2023qlora,he2022unifiedpeft},
but their objective is different: rather than only fitting target-task supervision, they aim to preserve or distill the behavioral effect induced by demonstrations.
DIVE follows this line of work, but studies a regime that is largely absent from prior multimodal steering experiments: long-form generation, where the intervention must remain reliable over many decoding steps and where errors in both content selection and sequence termination accumulate over time.

\paragraph{Long-form generation and salience-aware supervision.}
Autoregressive sequence models are trained under teacher forcing but decoded from their own histories, leading to exposure bias and compounding errors over long horizons
\citep{bengio2015scheduled,ranzato2016sequence}.
Prior work has also analyzed length bias, beam-search degradation, and degeneration in neural generation
\citep{wiseman2016beamsearch,murray2018length,cohen2019beam,meister2020beam,holtzman2020curious,welleck2020unlikelihood}.
These issues are particularly relevant to distilled latent steering: a fixed intervention estimated from teacher-forced states may become unreliable when applied to states visited during free-running student decoding.
In long-form medical generation, the imbalance is not only temporal but also semantic. 
The EOS event appears only once per report, and pathology-related finding mentions occupy only a small fraction of the sequence. 
Uniform token-level distillation can therefore be dominated by generic continuation tokens, while underrepresenting rare but high-impact decisions about which diagnostic findings to preserve and when to stop.
DIVE addresses this supervision-density imbalance during distillation through salience-aware supervision, together with state-conditioned steering that adapts the injected signal to the student's current decoding state.

\paragraph{Medical report generation and clinical evaluation.}
Medical report generation is a long-form multimodal task in which a model generates radiology findings from medical images.
It has been studied on datasets such as IU X-Ray, ChestX-ray8, CheXpert, CheXpert Plus, and MIMIC-CXR
\citep{demnerfushman2016iuxray,wang2017chestxray8,irvin2019chexpert,chambon2024chexpertplus,johnson2019mimic}.
Prior work has developed retrieval-augmented, memory-based, transformer-based, contrastive, longitudinal, and multiview report-generation models
\citep{jing2018medicalreports,li2018hybrid,chen2020r2gen,miura2021improving,yan2021weakly,wang2022r2gencmn,nicolson2024longitudinal,liu2025mlrg}.
Recent medical vision-language models further broaden the setting to few-shot and generalist biomedical assistants
\citep{li2023llavamed,moor2023medflamingo,tu2024generalist,zambrano2025llavarad}.
Because lexical overlap alone does not determine clinical correctness, evaluation increasingly relies on clinical-label, entity-relation, and error-oriented metrics such as CheXbert, RadGraph, and GREEN
\citep{smit2020chexbert,jain2021radgraph,yu2023evaluating,sloan2024review,ostmeier2024green}.
We use medical report generation as a stress test for long-form multimodal steering: the output is long enough for decoding drift to accumulate, and continuing beyond visually supported content can introduce unsupported clinical statements.

\begin{figure}[t]
    \centering
    \includegraphics[width=1.0\linewidth]{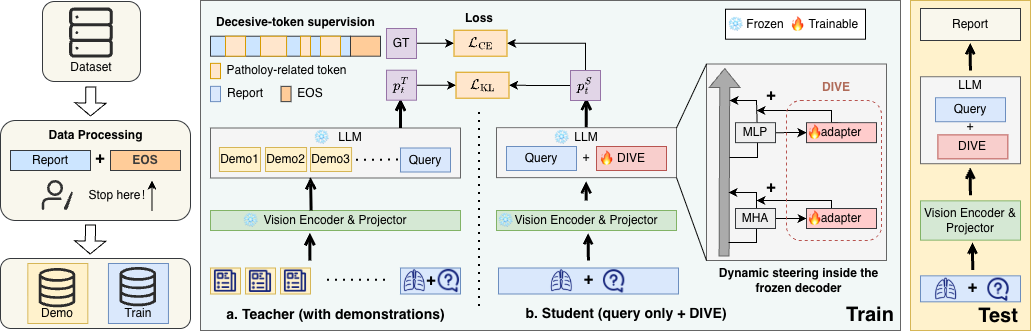}
    \caption{
\textbf{Training DIVE with decisive-token supervision and dynamic steering.}
A demonstration-augmented teacher provides cached token-level supervision for a query-only student.
DIVE combines \emph{decisive-token supervision}, which upweights pathology-related tokens and EOS in the cross-entropy loss, with \emph{dynamic steering}, which injects state-conditioned residuals into the frozen decoder through lightweight MHA/MLP adapters.
Training uses weighted token-level CE and top-$K$ KL distillation.
At inference time, demonstrations and teacher caches are removed, and only the learned adapters are used.
}
    \label{fig:dive_training}
\end{figure}

\section{DIVE: Dynamic In-context Vector Distillation with Decisive-token Supervision}

We propose DIVE, short for Dynamic In-context VEctor Distillation, a frozen-backbone framework for long-form multimodal generation. 
A demonstration-augmented teacher provides token-level supervision for a query-only student, while the base vision-language model remains frozen. 
DIVE combines two complementary mechanisms: state-conditioned dynamic steering, which adapts the injected intervention to the student's current decoding state, and decisive-token supervision, which upweights pathology-related target spans and the EOS event during distillation.
Implementation details, including norm-clipped injection, teacher-logit caching, and decoding configuration, are provided in Appendix~\ref{app:method_details}.

\subsection{Distilled Long-form Steering Setup}

Let $f_{\theta}$ denote a frozen vision-language model. 
Given a query image $x$ and instruction prompt $p$, the model generates a report
$y=(y_1,\ldots,y_T)$, where $y_T=\texttt{EOS}$ denotes the end-of-sequence token. 
In standard in-context learning, demonstrations $\mathcal{D}=\{d_i\}_{i=1}^{N}$ are prepended to the query input. 
DIVE distills this demonstration-augmented behavior into lightweight adapter parameters $\phi$, so that the student receives only $[p;x]$ during training and inference:
\[
p_t^S = f_{\theta,\phi}(y_t \mid p,x,y_{<t}).
\]
The teacher receives the demonstration-augmented input and is evaluated under teacher forcing:
\[
p_t^T = f_{\theta}(y_t \mid \mathcal{D},p,x,y_{<t}).
\]
Only the adapter parameters $\phi$ are optimized; all backbone parameters $\theta$ remain frozen.

\subsection{State-conditioned Dynamic Steering}

Static ICV-style steering applies a fixed residual vector to hidden states,
\[
\tilde{h}_{t}^{(l,b)} = h_{t}^{(l,b)} + v^{(l,b)},
\]
where $l$ indexes decoder layers and $b\in\{\mathrm{MHA},\mathrm{MLP}\}$ indexes Transformer branches. 
This open-loop intervention is brittle in long-form generation because the same residual is used even when autoregressive decoding drifts away from the teacher-forced trajectory.

DIVE replaces the fixed vector with a state-conditioned residual:
\[
\tilde{h}_{t}^{(l,b)} = h_{t}^{(l,b)} + \delta_{t}^{(l,b)}, 
\qquad 
\delta_{t}^{(l,b)} = g_{\phi}^{(l,b)}(h_{t}^{(l,b)}).
\]
In our implementation, $g_{\phi}^{(l,b)}$ is a bottleneck MLP adapter,
\[
\delta_{t}^{(l,b)}
=
W_{\mathrm{up}}^{(l,b)}
\sigma\!\left(
W_{\mathrm{down}}^{(l,b)} h_{t}^{(l,b)}
\right),
\]
where $W_{\mathrm{down}}^{(l,b)}\in\mathbb{R}^{r\times d}$,
$W_{\mathrm{up}}^{(l,b)}\in\mathbb{R}^{d\times r}$, and $r\ll d$. 
Because the residual is computed from the hidden state actually visited at each decoding step, the intervention can adapt to off-trajectory states during long-form generation.

\subsection{Decisive-token Distillation Objective}

% Long-form token-level distillation provides dense supervision for high-frequency template tokens, but much weaker effective supervision for tokens that determine clinical content and sequence boundaries. In general, decisive tokens are tokens that (a) appear with low frequency in the corpus, and (b) carry high semantic load for downstream evaluation. We instantiate this with EOS (low frequency by definition) and pathology mentions (low frequency from CheXpert-derived lexicon).
% In MRG, we treat two target-side token classes as decisive: pathology-related tokens that express clinically relevant findings, and the EOS event that marks the report boundary.
% To identify pathology-related tokens during training, we construct a CheXpert-guided phrase vocabulary covering common expressions of the 14 finding categories. 
% For each training report, we activate phrase sets corresponding to its finding labels and mark matched target-side tokens as pathology-related.
% This mask is used only to weight the supervised loss during adapter distillation; at inference time, DIVE uses no CheXpert labels, reference reports, pathology masks, teacher logits, or demonstrations.

Long-form token-level distillation provides dense supervision for high-frequency template tokens but much weaker effective supervision for the tokens that actually determine output quality. We refer to the latter as decisive tokens: target positions that (a) are sparse, either in absolute frequency or in their occurrence within a single output, and (b) disproportionately determine task-relevant content rather than surface fluency. Standard cross-entropy weights all positions equally and therefore allocates most gradient signal to abundant template tokens, leaving decisive positions underrepresented in the effective objective.
In MRG, this notion has two natural instances. Pathology-related tokens express the diagnostic findings that determine the clinical content of the report, and the EOS event determines whether the report terminates at the appropriate boundary. Both are sparse — pathology mentions occupy only a small fraction of each report, and EOS occupies a single position per sequence — yet both carry far more weight than the template phrasing that surrounds them.
To identify pathology-related tokens during training, we construct a CheXpert-guided phrase vocabulary covering common surface expressions of the 14 finding categories (full lexicon in Appendix C). For each training report, we activate phrase sets corresponding to its CheXpert finding labels and mark matched target-side tokens as pathology-related, yielding a per-example pathology mask. The EOS mask is defined trivially as the indicator of the EOS position. Both masks are used only to weight the distillation loss during training; at inference time, DIVE uses no CheXpert labels, reference reports, pathology masks, teacher logits, or demonstrations.
Let $m_t^{\mathrm{path}}\in\{0,1\}$ denote this pathology-token mask, and let $m_t^{\mathrm{EOS}}=\mathbb{I}[y_t=\texttt{EOS}]$ denote the EOS mask.
DIVE then upweights the cross-entropy contribution of both decisive token types:
\[
\mathcal{L}_{\mathrm{CE}}
=
\frac{
\sum_{t=1}^{T} w_t \, \mathrm{CE}(p_t^S, y_t)
}{
\sum_{t=1}^{T} w_t
},
\qquad
w_t =
\begin{cases}
\omega_{\mathrm{EOS}}, & m_t^{\mathrm{EOS}}=1,\\
\omega_{\mathrm{path}}, & m_t^{\mathrm{path}}=1,\\
1, & \text{otherwise}.
\end{cases}
\]
This changes the training objective rather than applying a decoding-time length penalty or forced stopping rule. 
It increases supervision on the sparse tokens responsible for diagnostic content and termination while keeping ordinary template and grammatical tokens in the objective.

To transfer the demonstration-induced token preferences of the teacher, we also minimize a top-$K$ KL distillation loss:
\[
\mathcal{L}_{\mathrm{KL}}
=
\frac{1}{T}
\sum_{t=1}^{T}
\mathrm{KL}
\left(
p^{T}_{t,K}
\,\Vert\,
p^{S}_{t,K}
\right),
\]
where $p^{T}_{t,K}$ and $p^{S}_{t,K}$ are teacher and student distributions restricted to the teacher's top-$K$ token support. 
The final objective is
\[
\mathcal{L}
=
\alpha \mathcal{L}_{\mathrm{KL}}
+
(1-\alpha)\mathcal{L}_{\mathrm{CE}}.
\]
At inference time, DIVE discards demonstrations, teacher caches, and the pathology mask used only as a training-time domain prior, relying solely on the query image, instruction prompt, frozen backbone, and learned dynamic adapters.

\section{Experiment}

\subsection{Settings and Implementation Details}

\textbf{Model and dataset.}
We evaluate DIVE on two medical vision-language backbones, QoQ-Med3-VL-8B~\citep{dai2025qoqmed} and LLaVA-Med v1.5 Mistral-7B~\citep{li2023llavamed}, and on two chest X-ray report generation benchmarks, MIMIC-CXR~\citep{johnson2019mimic} and CheXpert Plus~\citep{chambon2024chexpertplus}. After preprocessing and deduplication, MIMIC-CXR contains 214,996 valid training samples and 3,087 test samples, while CheXpert Plus contains 12,816 valid training samples and 3,663 test samples. 
For each dataset, we randomly sample 1,000 training instances for adapter distillation and 2,000 instances as the demonstration pool. Final results are reported on the full test set for each dataset-backbone pair.

\textbf{Distillation setting.}
For each query, we construct text-only in-context demonstrations to obtain teacher logits. 
QoQ-Med uses 32-shot demonstrations, while LLaVA-Med uses 8-shot demonstrations due to its effective context length limitation. 
Dynamic task vectors are implemented as per-layer bottleneck MLP injectors with rank 16. We train them using a weighted combination of top-K KL distillation and cross-entropy on answer tokens, with pathology-related target spans and EOS explicitly upweighted in the cross-entropy term. Full implementation details, including optimization, temperature scheduling, quantization, and hardware configuration, are deferred to the appendix.

\textbf{Evaluation metrics.}
We evaluate generated reports with standard text-generation metrics, including BLEU-1/4, ROUGE-L, CIDEr, and BERTScore
\citep{papineni2002bleu,lin2004rouge,vedantam2015cider,zhang2020bertscore}.
To assess clinical content beyond lexical overlap, we also report CheXbert-14 F1 and RadGraph F1
\citep{smit2020chexbert,jain2021radgraph}.
Following recent analyses of radiology report generation, we treat these automatic metrics as clinical proxies rather than clinical validation
\citep{yu2023evaluating,sloan2024review,ostmeier2024green}.

\subsection{Main Results}

\begin{table}[t]
\centering
\footnotesize
\setlength{\tabcolsep}{3.5pt}
\caption{
Performance (\%) comparison on the MIMIC-CXR and CheXpert Plus test sets under different medical VLM backbones.
BERTScore is reported as F1.
Bold and underline indicate the best and second-best results within each dataset--backbone block.
Significance markers indicate uncorrected one-sided paired tests against the strongest non-DIVE baseline under the same dataset and backbone:
$^{***}p<0.001$, $^{**}p<0.01$, $^{*}p<0.05$.
}
\label{tab:main_results}
\resizebox{\linewidth}{!}{%
\begin{tabular}{llccccccc}
\toprule
Backbone & Model & BLEU-1 & BLEU-4 & ROUGE-L & CIDEr & BERTScore & F1-chexbert-14 & F1-radgraph \\
\midrule
\multicolumn{9}{l}{\textit{MIMIC-CXR}} \\
\midrule
QoQ-Med
& Zero-shot   & 17.59 & 1.30 & 14.60 & 0.62 & 84.19 & 36.49 & 8.64 \\
& 32-shot ICL & 22.23 & 2.87 & 17.37 & 1.08 & 85.57 & \textbf{45.97} & 15.79 \\
& QLoRA       & 19.33 & 3.89 & 18.17 & \textbf{7.38} & \textbf{86.51} & 34.09 & 14.78 \\
& LIVE        & \underline{24.19} & \underline{4.39} & \underline{19.05} & 2.78 & 85.75 & 34.30 & \underline{18.60} \\
& DIVE (Ours) & \textbf{29.69}$^{***}$ & \textbf{5.33}$^{***}$ & \textbf{20.18}$^{***}$ & \underline{7.15} & \underline{86.45} & \underline{37.91} & \textbf{19.00}$^{**}$ \\
\cmidrule(lr){1-9}
LLaVA-Med
& Zero-shot   & 18.55 & 1.60 & 14.89 & 1.16 & 84.09 & 34.22 & 11.26 \\
& 8-shot ICL  & \underline{25.19} & 3.22 & 18.06 & 2.94 & 85.68 & \textbf{41.86} & 16.21 \\
& QLoRA       & 18.07 & \underline{3.31} & \underline{18.48} & \underline{5.56} & \underline{86.76} & 25.87 & \underline{16.94} \\
& LIVE        & 21.42 & 2.26 & 16.37 & 2.20 & 84.39 & 16.03 & 15.75 \\
& DIVE (Ours) & \textbf{30.93}$^{***}$ & \textbf{4.79}$^{***}$ & \textbf{20.49}$^{***}$ & \textbf{5.85} & \textbf{87.10}$^{***}$ & \underline{37.42} & \textbf{18.06}$^{***}$ \\
\midrule
\multicolumn{9}{l}{\textit{CheXpert Plus}} \\
\midrule
QoQ-Med
& Zero-shot   & 17.87 & 1.62 & 15.46 & 0.69 & 84.61 & 35.33 & 10.30 \\
& 32-shot ICL & \textbf{21.06} & 2.54 & 15.87 & 2.77 & \underline{84.87} & 18.43 & 13.09 \\
& QLoRA       & 15.98 & 3.02 & 16.30 & 1.07 & 83.18 & \underline{39.92} & 13.81 \\
& LIVE        & 20.06 & \underline{3.23} & \underline{17.74} & \underline{3.98} & 84.60 & 38.61 & \underline{15.00} \\
& DIVE (Ours) & \underline{20.37} & \textbf{3.93}$^{***}$ & \textbf{18.76}$^{***}$ & \textbf{4.65}$^{*}$ & \textbf{85.05}$^{***}$ & \textbf{41.28} & \textbf{17.18}$^{***}$ \\
\cmidrule(lr){1-9}
LLaVA-Med
& Zero-shot   & 11.69 & 0.48 & 10.97 & 1.44 & 83.97 & 24.62 & 4.84 \\
& 8-shot ICL  & \underline{23.04} & 2.48 & 15.55 & \underline{3.49} & \underline{85.12} & 32.06 & 10.81 \\
& QLoRA       & 18.83 & \underline{3.50} & \underline{17.21} & 2.02 & 83.92 & \underline{40.15} & 13.75 \\
& LIVE        & 22.52 & 2.03 & 16.85 & 2.24 & 85.10 & 28.36 & \underline{14.47} \\
& DIVE (Ours) & \textbf{27.79}$^{***}$ & \textbf{5.31}$^{***}$ & \textbf{18.99}$^{***}$ & \textbf{7.49}$^{***}$ & \textbf{85.86}$^{***}$ & \textbf{43.57}$^{***}$ & \textbf{16.88}$^{***}$ \\
\bottomrule
\end{tabular}%
}
\end{table}

Table~\ref{tab:main_results} compares DIVE with zero-shot prompting, in-context prompting, QLoRA, and LIVE across two datasets and two medical VLM backbones.
DIVE shows the most consistent gains over the direct static-steering baseline, outperforming LIVE across all reported metrics in every dataset--backbone setting.
The improvements are especially clear on lexical quality and structured clinical fidelity: DIVE achieves the best BLEU-4, ROUGE-L, and RadGraph F1 in all four settings.

At the same time, DIVE does not uniformly maximize every clinical proxy.
On MIMIC-CXR, ICL obtains the highest CheXbert-14 F1, while DIVE remains second-best and achieves the strongest RadGraph F1.
This suggests that DIVE improves the trade-off between report-level quality, structured entity--relation fidelity, and inference efficiency, rather than simply optimizing a single label-level clinical metric.

\subsection{Component Ablation}

\begin{table*}[t]
\centering
\footnotesize
\setlength{\tabcolsep}{4pt}
\caption{Cumulative ablation of LLaVA-Med on CheXpert Plus and MIMIC-CXR. Starting from the zero-shot baseline, each subsequent row adds one component.}
\label{tab:llava_ablation}
\resizebox{\textwidth}{!}{%
\begin{tabular}{lccccccc}
\toprule
Setting & BLEU-1 & BLEU-4 & ROUGE-L & CIDEr & BERTScore & F1-CheXbert & F1-RadGraph \\
\midrule
\multicolumn{8}{l}{\textit{CheXpert Plus}} \\
Zero-shot & 11.69 & 0.48 & 10.97 & 1.44 & 83.97 & 24.62 & 4.84 \\
+ Dynamic TV & 19.75 & 3.89 & 18.49 & 6.72 & \textbf{86.09} & 23.30 & 15.40 \\
+ Pathology-token supervision & 21.05 & 4.15 & 18.38 & 6.70 & 85.89 & 33.53 & 15.15 \\
+ EOS ($w=1$) & 23.00 & 4.73 & 18.79 & \textbf{7.59} & 85.97 & 35.38 & 15.39 \\
+ EOS upweight ($w=5$) & \textbf{27.79} & \textbf{5.31} & \textbf{18.99} & 7.49 & 85.86 & \textbf{43.57} & \textbf{16.88} \\
\midrule
\multicolumn{8}{l}{\textit{MIMIC-CXR}} \\
Zero-shot & 18.55 & 1.60 & 14.89 & 1.16 & 84.09 & 34.22 & 11.26 \\
+ Dynamic TV & 22.16 & 3.07 & 18.66 & 1.62 & 85.53 & 28.80 & 19.49 \\
+ Pathology-token supervision & 24.42 & 3.89 & 19.15 & 2.66 & 86.02 & 33.70 & \textbf{22.00} \\
+ EOS ($w=1$) & 22.88 & 3.17 & 18.00 & 1.84 & 85.99 & 32.85 & 17.90 \\
+ EOS upweight ($w=5$) & \textbf{30.93} & \textbf{4.79} & \textbf{20.49} & \textbf{5.85} & \textbf{87.10} & \textbf{37.42} & 18.06 \\
\bottomrule
\end{tabular}%
}
\end{table*}

Table~\ref{tab:llava_ablation} isolates the contribution of each DIVE component with the LLaVA-Med backbone.
Dynamic task-vector injection gives a strong first-stage improvement over zero-shot generation, showing that state-conditioned steering can distill useful demonstration-induced behavior without retaining examples at inference time.
Pathology-token supervision provides complementary clinical gains, particularly improving CheXbert F1 on CheXpert Plus and RadGraph F1 on MIMIC-CXR.

EOS supervision changes the balance between report quality and clinical coverage.
EOS upweighting yields the strongest overall performance on CheXpert Plus and the best text-generation metrics and CheXbert F1 on MIMIC-CXR, while the pathology-weighted variant retains the highest MIMIC-CXR RadGraph F1.
Thus, the final DIVE variant should be interpreted as optimizing a balanced quality--clinical-fidelity--stopping trade-off, rather than uniformly maximizing every proxy metric.

\subsection{Stopping Behavior and EOS Calibration}
\noindent
\begin{table*}[t]
\centering
\footnotesize
\setlength{\tabcolsep}{4.5pt}
\caption{
Length-control behavior of LLaVA-Med ablations. 
For each example, $\Delta\mathrm{Len}$ is the generated-reference word-count difference. 
Avg. $\Delta$Len and MAE report its mean and mean absolute error. 
U/O/P reports the percentage of examples with $\Delta\mathrm{Len}<-5$, $\Delta\mathrm{Len}>5$, and $|\Delta\mathrm{Len}|\leq5$, respectively.
}
\label{tab:llava_length_control}
\resizebox{\textwidth}{!}{%
\begin{tabular}{lcccccccc}
\toprule
\multirow{2}{*}{Model}
& \multicolumn{4}{c}{CheXpert Plus}
& \multicolumn{4}{c}{MIMIC-CXR} \\
\cmidrule(lr){2-5} \cmidrule(lr){6-9}
& Gen Len & Avg. $\Delta$Len & MAE & U/O/P (\%)
& Gen Len & Avg. $\Delta$Len & MAE & U/O/P (\%) \\
\midrule
Zero-shot baseline
& 29.78 & -15.08 & 20.26 & 57.77 / 18.10 / 24.13
& 78.10 & +26.79 & 29.93 & 7.52 / 85.88 / 6.61 \\

+ Dynamic TV
& 28.38 & -16.48 & 20.22 & 62.38 / 14.11 / 23.51
& 83.24 & +31.93 & 34.39 & 5.77 / 87.24 / 7.00 \\

+ Pathology-token supervision
& 31.69 & -13.17 & 19.42 & 54.03 / 19.44 / 26.54
& 74.76 & +23.45 & 27.36 & 9.10 / 81.60 / 9.30 \\

+ EOS ($w=1$)
& 33.85 & -11.01 & \textbf{18.74} & 49.49 / 23.04 / 27.46
& 75.00 & +23.69 & 27.70 & 9.14 / 82.86 / 8.00 \\

+ EOS upweight ($w=5$)
& 45.13 & \textbf{+0.27} & 18.99 & \textbf{30.49} / 39.99 / \textbf{29.51}
& 46.31 & \textbf{-5.00} & \textbf{17.23} & 44.80 / \textbf{31.84} / \textbf{23.36} \\
\bottomrule
\end{tabular}%
}
\end{table*}

\begin{figure*}[t]
\centering
\includegraphics[width=0.8\textwidth]{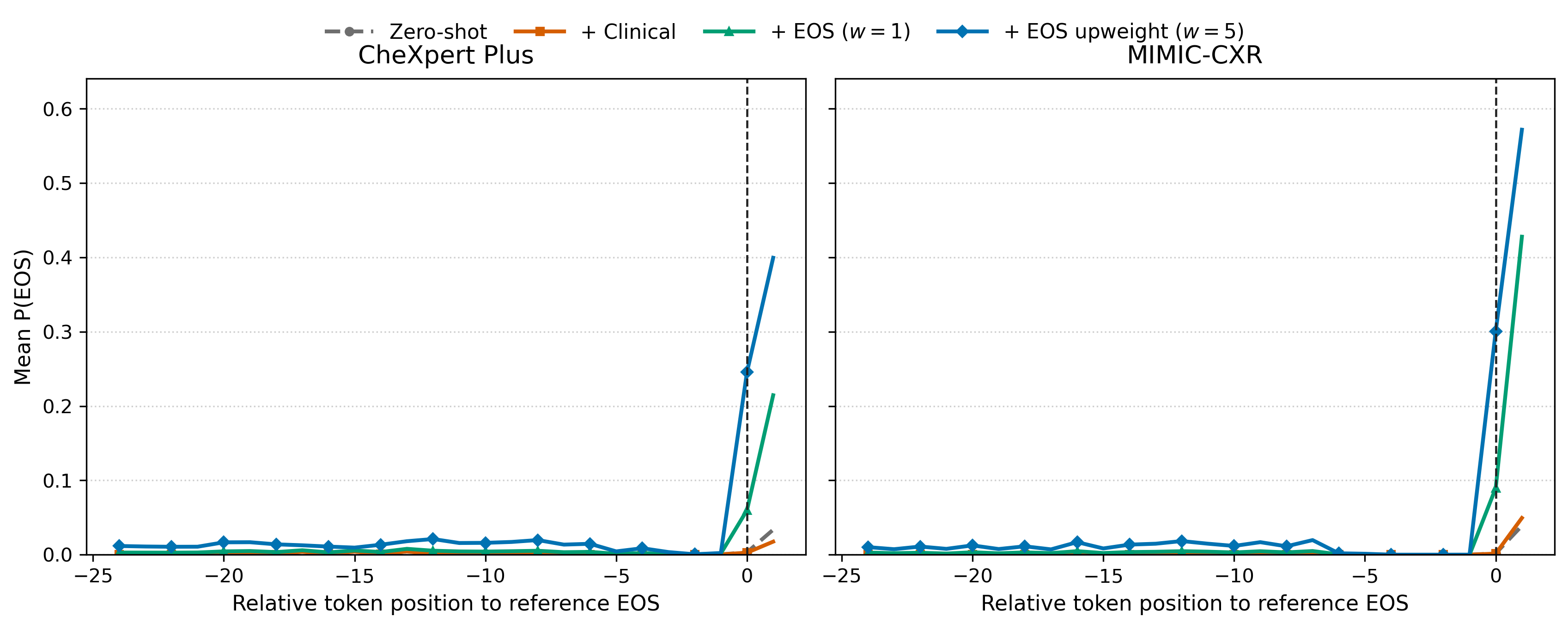}
\caption{
EOS probability around the reference report boundary on CheXpert Plus and MIMIC-CXR.
Position 0 denotes the reference EOS token.
Variants without EOS supervision assign little probability mass near the boundary, whereas EOS supervision produces a localized EOS peak, further strengthened by EOS upweighting.
}
\label{fig:llava_eos_boundary}
\end{figure*}

Table~\ref{tab:llava_length_control} evaluates termination behavior across LLaVA-Med ablations.
The two datasets exhibit different initial length biases: zero-shot LLaVA-Med tends to under-generate on CheXpert Plus, but substantially over-generates on MIMIC-CXR.
Dynamic task-vector injection and pathology-token supervision improve content modeling, yet they do not by themselves reliably place probability mass on the correct stopping boundary.

EOS supervision specifically targets this missing termination signal.
Adding EOS with unit weight reduces the under-generation bias on CheXpert Plus, while EOS upweighting further shifts the stopping distribution: it brings the mean length difference close to zero on CheXpert Plus and markedly reduces over-generation on MIMIC-CXR.
Figure~\ref{fig:llava_eos_boundary} supports this interpretation at the token level. Variants without EOS supervision assign little probability mass near the reference boundary, whereas EOS-supervised variants produce a localized EOS probability peak around the reference EOS position.
This indicates that the observed stopping improvements arise from explicit EOS supervision, rather than from generic content steering alone.

Importantly, these results show that stopping is a calibration trade-off rather than a simple length-minimization problem.
Stronger EOS weighting suppresses late continuation, but can also increase under-generation, especially on MIMIC-CXR.
Therefore, the objective should be interpreted as balancing under-generation, over-generation, and proper stopping, rather than simply encouraging shorter reports.

\subsection{Effective Supervision Mass of Decisive Tokens}
\label{sec:token-supervision-mass}

\begin{wrapfigure}{r}{0.48\linewidth}
    \vspace{-1.0em}
    \centering
    \includegraphics[width=\linewidth]{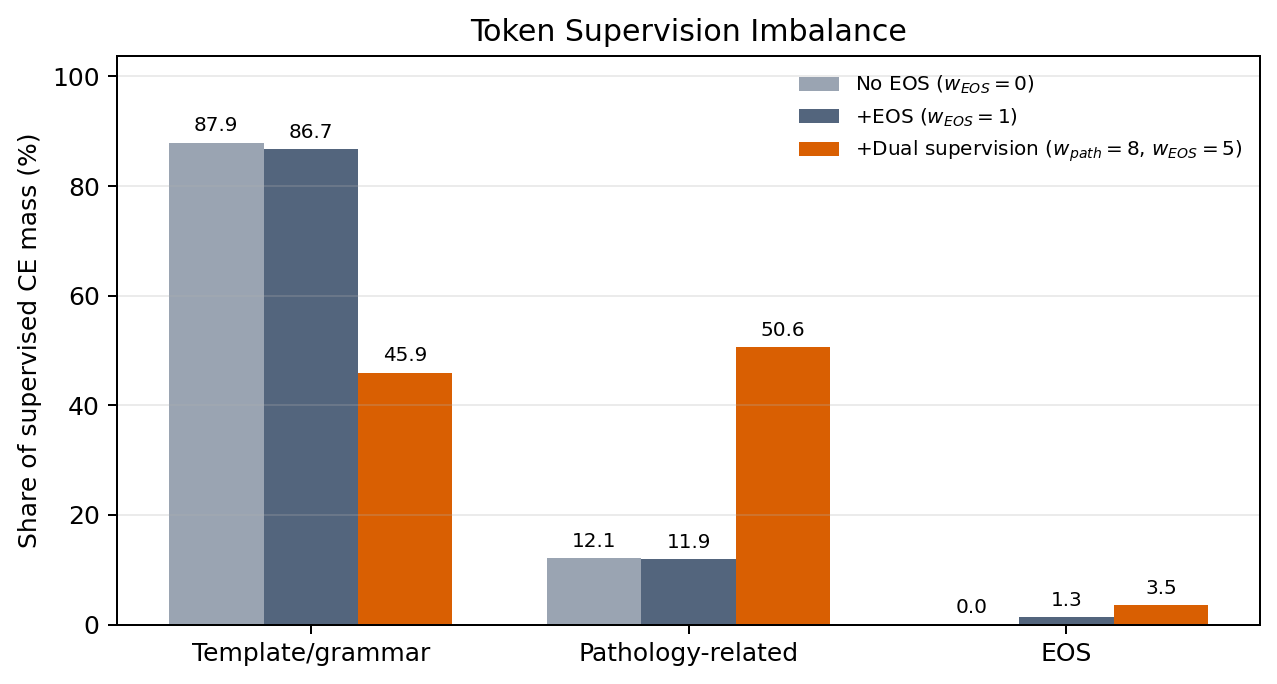}
    \caption{
Effective CE supervision mass on the LLaVA-Med CheXpert Plus training split.
DIVE's dual supervision ($w_{\mathrm{path}}=8$, $w_{\mathrm{EOS}}=5$) shifts supervision from high-frequency template/grammar tokens toward pathology-related tokens and EOS.
}
    \label{fig:token-supervision-mass}
    \vspace{-1.0em}
\end{wrapfigure}

To quantify the supervision imbalance behind the ablation results, we measure the effective CE mass assigned to template/grammar tokens, pathology-related tokens, and EOS on the LLaVA-Med CheXpert Plus training split.
For category $c$, we define
$M(c)=\sum_{t:y_t\in c}w_t$, where $w_t$ is the CE weight applied to target token $y_t$.

As shown in Figure~\ref{fig:token-supervision-mass}, standard supervision is dominated by high-frequency template and grammatical tokens, while EOS receives little or no effective mass.
DIVE's dual supervision reallocates training signal toward pathology-related tokens and EOS without removing ordinary language-modeling supervision.
This token-level redistribution explains why pathology-token supervision and EOS upweighting provide complementary gains in the ablations and improve stopping calibration.

\subsection{Forward-Pass Efficiency Analysis}
\label{sec:forward-pass-efficiency}

We further compare DIVE and vanilla ICL under a single forward operation, with FLOPs and runtime normalized by zero-shot inference.
Because DIVE uses the same input context as zero-shot inference, it preserves the same FLOPs cost and incurs only a small runtime overhead from task-vector injection.
In contrast, vanilla ICL expands the context with demonstrations, causing both FLOPs and runtime to increase steadily as the number of shots grows.

\begin{figure}[t]
    \centering
    \includegraphics[width=0.82\linewidth]{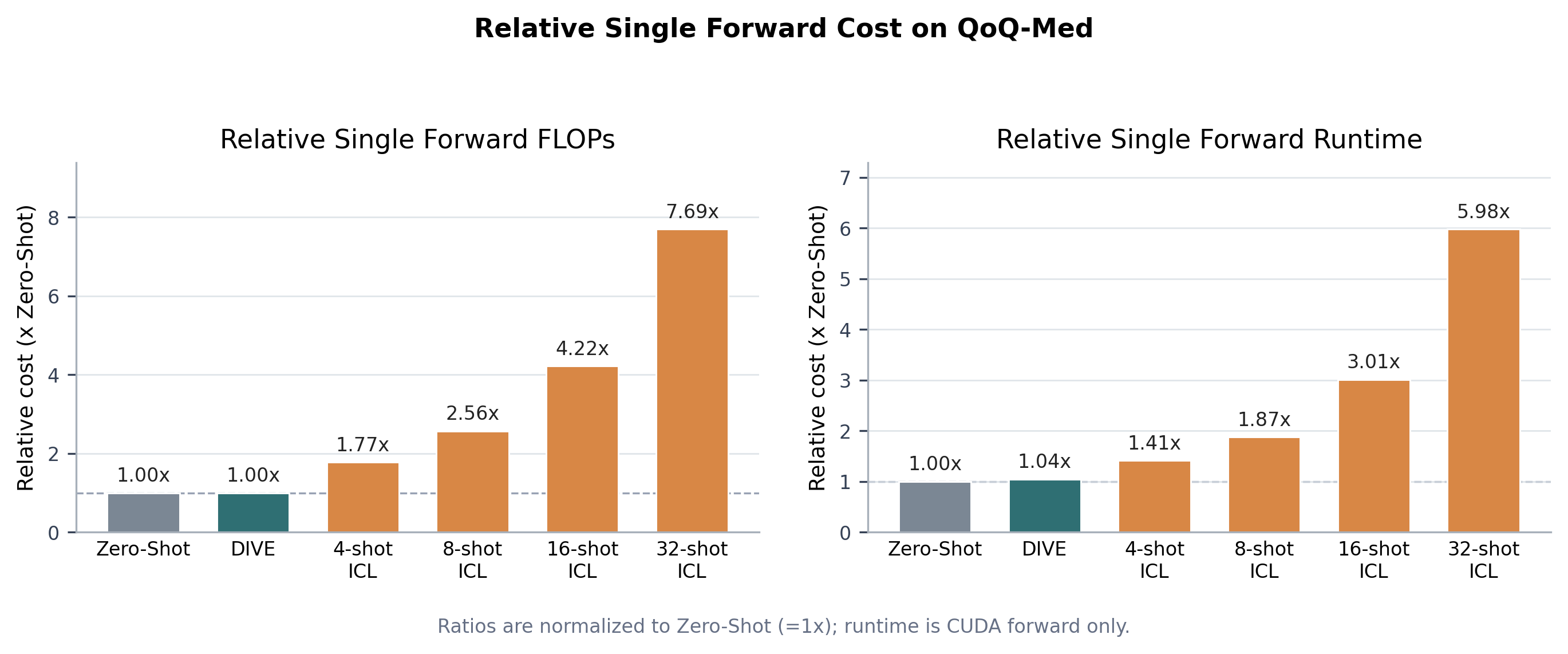}
    \caption{
    Relative single-forward cost on QoQ-Med.
    FLOPs and runtime are normalized by zero-shot inference.
    DIVE maintains nearly zero-shot cost, while vanilla ICL incurs increasing overhead as the number of demonstrations grows.
    }
    \label{fig:forward-cost-qoq-ratio}
\end{figure}

Figure~\ref{fig:forward-cost-qoq-ratio} shows that DIVE preserves the practical efficiency advantage of distilled adaptation: demonstrations are used to construct the training signal, but they are removed at inference time.
This is important for long-form medical report generation, where adding demonstrations to the context increases the cost of each decoding step and can become expensive as the number of shots grows.

\section{Limitations and Conclusion}

We studied long-form medical report generation as a challenging setting for distilled multimodal in-context steering. 
Our analysis shows that transferring hidden-space steering from short-form tasks to long-form reports exposes two coupled failures: the steering signal must remain reliable under autoregressive trajectory drift, and the training objective must represent sparse but high-impact supervision targets. 
In radiology reports, these targets include both clinically salient pathology-related spans, which determine what content should be preserved, and the EOS boundary, which determines when generation should stop.

We proposed DIVE, a frozen-backbone distillation framework that combines state-conditioned dynamic steering with dual salience-aware supervision. 
The dynamic injector adapts the intervention to the student's current decoding state, while the salience-aware objective upweights CheXpert-guided clinical spans and the EOS event during distillation. 
Across MIMIC-CXR and CheXpert Plus with two medical vision-language backbones, DIVE consistently ranks among the strongest methods across report-generation, clinical-fidelity, and stopping metrics. 
The gains are especially pronounced in structured clinical fidelity and termination calibration, indicating that the method improves the trade-off between what to generate and when to stop, rather than merely increasing output length or applying a decoding-time heuristic.

At the same time, DIVE should not be interpreted as a clinically validated report-generation system. 
Automatic metrics such as CheXbert and RadGraph are useful clinical proxies, but they do not replace expert radiologist evaluation. 
Our results also show that improving one aspect of clinical fidelity does not uniformly optimize every proxy metric: stronger clinical-span or EOS weighting can shift the balance among recall, unsupported findings, and under-generation. 
Future work should evaluate DIVE with expert clinical review, stronger factuality objectives, pathology-specific error analysis, and broader long-form multimodal tasks beyond chest X-ray report generation.

\clearpage
\bibliographystyle{plainnat}
\bibliography{references}

%%%%%%%%%%%%%%%%%%%%%%%%%%%%%%%%%%%%%%%%%%%%%%%%%%%%%%%%%%%%
\clearpage
\appendix

%%%%%%%%%%%%%%%%%%%%%%%%%%%%%%%%%%%%%%%%%%%%%%%%%%%%%%%%%%%%

\section{Implementation Details}

\subsection{Datasets and Preprocessing}
We conduct experiments on two chest X-ray report generation benchmarks: MIMIC-CXR-JPG~\citep{johnson2019mimic} and CheXpert Plus~~\citep{chambon2024chexpertplus}. Both datasets contain chest radiographs paired with corresponding radiology report findings. We convert all samples into a unified chat-style format, where the query case contains an image and the target output is the findings section. We remove samples with missing image paths or empty target reports, and deduplicate the data by study identifier when applicable.

\begin{table}[H]
\centering
\caption{Dataset statistics and experimental subsets.}
\label{tab:dataset_setup}
\small
\begin{tabular}{lccccc}
\toprule
Dataset & Train & Adapter Train & Demo Pool & Validation & Test \\
\midrule
MIMIC-CXR-JPG & 214,996 & 1,000 & 2,000 & 200 & 3,087 \\
CheXpert Plus & 12,816 & 1,000 & 2,000 & 200 & 3,663 \\
\bottomrule
\end{tabular}
\end{table}
\FloatBarrier

For all methods, the target output is the report findings section. Generated reports are capped at 128 new tokens. In-context demonstrations are text-only findings reports sampled from the training demonstration pool; only the query case includes an image.

\subsection{Model Backbones}
We evaluate our method on two medical vision-language backbones, following recent work on medical and radiology vision-language models~\citep{li2023llavamed,moor2023medflamingo,tu2024generalist,zambrano2025llavarad}.

\begin{table}[H]
\centering
\caption{Backbones used in our experiments.}
\label{tab:backbones}
\small
\begin{tabular}{lccc}
\toprule
Backbone & Base Model & Context Setting & Demonstrations \\
\midrule
QoQ-Med & QoQ-Med3-VL-8B & 4-bit quantized VLM & 32-shot ICL \\
LLaVA-Med & LLaVA-Med v1.5 Mistral-7B & 4-bit quantized VLM & 8-shot ICL \\
\bottomrule
\end{tabular}
\end{table}
\FloatBarrier

QoQ-Med supports a substantially larger context window, so it can accommodate high-shot textual demonstrations. LLaVA-Med has a much shorter effective context after image-token expansion. Because each query image is internally expanded into visual patch embeddings, longer full-demonstration prompts can truncate the teacher-forced answer region. Therefore, for LLaVA-Med we restrict all ICL-based experiments to 8-shot demonstrations.

\subsection{Baseline Details}
We compare DIVE with the following baselines:

\begin{itemize}
    \item \textbf{Zero-shot.} The frozen backbone receives only the query image and the report-generation instruction, without demonstrations or adapter injection.

    \item \textbf{ICL.} The frozen backbone receives text-only demonstration findings reports before the query image. QoQ-Med uses the high-shot ICL setting, while LLaVA-Med uses 8-shot ICL due to its effective context limit.

    \item \textbf{QLoRA}~\citep{dettmers2023qlora}. A parameter-efficient supervised fine-tuning baseline. Low-rank adapters are trained on the same 1,000 training samples while the backbone remains quantized.

    \item \textbf{LIVE}~\citep{peng2024live}. A hidden-state shift baseline that learns layer-wise output shifts from an ICL teacher and applies them during single-image student generation.

    \item \textbf{DIVE.} Our method distills the ICL teacher into dynamic task-vector adapters. The student uses only the query image at inference time, while the learned adapters approximate the behavior induced by in-context demonstrations.
\end{itemize}
\FloatBarrier

\subsection{Training Hyperparameters}
For DIVE, we cache the top-200 teacher logits for each answer token. The distillation objective combines top-$k$ KL imitation and supervised cross-entropy:
\[
\mathcal{L}
=
\alpha \mathcal{L}_{\mathrm{KL}}
+
(1-\alpha)\mathcal{L}_{\mathrm{CE}},
\]

\begin{table}[H]
\centering
\footnotesize
\setlength{\tabcolsep}{4pt}
\caption{DIVE-specific training and distillation settings.}
\label{tab:chextxp_best_configs}
\small
\begin{tabular}{lcc}
\toprule
Setting & QoQ-Med & LLaVA-Med \\
\midrule
Backbone & QoQ-Med3-VL-8B & LLaVA-Med-v1.5-Mistral-7B \\
Trainable Parameters & 9.44M & 	8.39M \\
ICL demonstrations & 32 & 8 \\
Epochs & 5 & 5 \\
Learning rate & $1{\times}10^{-4}$ & $1{\times}10^{-4}$ \\
Distillation weight $\alpha$ & 0.8 & 0.8 \\
Temperature & 2.0  & 2.0  \\
Injection mode & all tokens & all tokens \\
Decode decay rate & 0.9 & 0.9 \\
EOS weight & 5 & 5 \\
Clinical finding CE weight & 3 & 8 \\
\bottomrule
\end{tabular}
\end{table}

\FloatBarrier

\subsection{Statistical Testing and Reproducibility}

For the main result table, significance markers are computed by comparing DIVE with the strongest non-DIVE baseline separately for each metric and backbone. The strongest baseline is selected according to the reported test-set score for that metric. We use paired $t$-tests over sample-level per-example scores on the MIMIC-CXR and CheXpert Plus, pairing systems by the same test example. The reported markers use uncorrected one-sided $p$-values for the alternative hypothesis $H_1$: DIVE obtains a higher score than the baseline. We use the thresholds $^{*}p<0.05$, $^{**}p<0.01$, and $^{***}p<0.001$.

\subsection{The Detailed Inference Speed Experiments}
\label{sec:detailed-inference-speed}

To complement the relative efficiency comparison, we further report the detailed
single-forward cost of DIVE and vanilla ICL on QoQ-Med. This experiment measures
the cost of one forward pass over the input context, excluding autoregressive
decoding length and evaluation-side overhead. FLOPs are estimated with an
attention-adjusted token proxy, and runtime is measured as CUDA forward latency.

\begin{table}[H]
\centering
\small
\caption{
Detailed single-forward inference cost on QoQ-Med.
Runtime is measured per forward pass, and relative costs are normalized by
zero-shot inference.
}
\label{tab:detailed-inference-speed}
\resizebox{\linewidth}{!}{
\begin{tabular}{lccccc}
\toprule
Method & Avg. Tokens & FLOPs (TFLOPs) & FLOPs Ratio & Runtime (ms) & Runtime Ratio \\
\midrule
Zero-Shot    & 413  & 6.709  & 1.00$\times$ & 373.06  & 1.00$\times$ \\
DIVE         & 413  & 6.709  & 1.00$\times$ & 388.83  & 1.04$\times$ \\
4-shot ICL   & 724  & 11.898 & 1.77$\times$ & 526.78  & 1.41$\times$ \\
8-shot ICL   & 1036 & 17.204 & 2.56$\times$ & 699.43  & 1.87$\times$ \\
16-shot ICL  & 1665 & 28.279 & 4.22$\times$ & 1122.48 & 3.01$\times$ \\
32-shot ICL  & 2913 & 51.607 & 7.69$\times$ & 2230.32 & 5.98$\times$ \\
\bottomrule
\end{tabular}
}
\end{table}

\subsection{Compute Resources and Inference Cost}
\label{sec:compute-resources}

All experiments were conducted on NVIDIA RTX A5000 GPUs with 24GB memory.
For DIVE training, the vision-language backbone is kept frozen and loaded in 4-bit precision, and only the task-vector modules are optimized.

To quantify practical inference cost, we additionally measure end-to-end generation latency and peak GPU memory on QoQ-Med and LLaVA-Med set.
The benchmark uses batch size 4, excludes model-loading time, and reports per-sample latency as batch wall-clock time divided by batch size.
Peak memory is measured with CUDA peak allocated memory during generation.
Unlike the single-forward analysis in Section~\ref{sec:forward-pass-efficiency}, this benchmark uses actual EOS termination and therefore reflects both computational cost and stopping behavior.

\begin{table}[H]
\centering
\small
\caption{
End-to-end generation latency and peak GPU memory on QoQ-Med and MIMIC-CXR set.
}
\label{tab:generation-latency-memory}
\begin{tabular}{lccc}
\toprule
Method & Gen. Tokens & Latency (s) & Peak Mem. (GB) \\
\midrule
Zero-shot   & 127.9 & 5.15 & 9.80 \\
DIVE        & 84.8  & 3.57 & 9.80 \\
4-shot ICL  & 127.2 & 5.52 & 10.38 \\
8-shot ICL  & 126.4 & 5.88 & 10.97 \\
16-shot ICL & 127.5 & 6.79 & 12.28 \\
32-shot ICL & 127.9 & 8.65 & 14.83 \\
\bottomrule
\end{tabular}
\end{table}

As shown in Table~\ref{tab:generation-latency-memory}, DIVE preserves the same prompt length and peak memory footprint as zero-shot inference, while reducing end-to-end generation latency due to earlier EOS termination.
In contrast, vanilla ICL increases both latency and memory as demonstrations are added to the context.
At 32 shots, ICL requires 8.65 seconds per sample and 14.83GB peak memory, compared with 3.57 seconds and 9.80GB for DIVE.

\subsection{Adapter Architecture and Injection Layers}
DIVE freezes the full vision-language backbone and trains only lightweight dynamic task-vector adapters. For each transformer layer and each branch, the adapter computes a low-rank hidden-state update:
\[
\Delta h_{\ell,b}
=
W^{\mathrm{up}}_{\ell,b}
\sigma
\left(
W^{\mathrm{down}}_{\ell,b} h_{\ell,b}
\right),
\]
where $b$ denotes either the self-attention branch or the MLP branch. We use rank-16 bottlenecks, and initialize the up-projection to zero so that the initial model behavior matches the frozen backbone.

\begin{table}[H]
\centering
\caption{Adapter architectures and injection locations.}
\label{tab:adapter_arch}
\small
\begin{tabular}{lccc}
\toprule
Method & Trainable Module & Injection Location & Trainable Scope \\
\midrule
QLoRA & Low-rank adapters & Attention and FFN projections & Selected LM modules \\
LIVE & Layer-wise vector shift & Decoder layer outputs & All decoder layers \\
DIVE & Dynamic task vectors & Self-attention and MLP outputs & All decoder layers \\
\bottomrule
\end{tabular}
\end{table}
\FloatBarrier

The DIVE residual update is added using a norm-preserving operation to stabilize generation. For LLaVA-Med, we additionally align teacher-forced logits with the expanded multimodal token positions before caching teacher distributions, since the raw text-token positions differ from the internal model positions after image-token expansion.

\subsection{Evaluation Metrics}
We evaluate generated reports with two groups of metrics:

\begin{itemize}
    \item \textbf{Text generation metrics.} We report BLEU-1, BLEU-4, ROUGE-L, CIDEr, and BERTScore-F1~\citep{papineni2002bleu,lin2004rouge,vedantam2015cider,zhang2020bertscore}.

    \item \textbf{Clinical correctness metrics.} We report micro-averaged F1 over 14 CheXbert labels and F1-RadGraph to evaluate clinical entity and relation correctness~\citep{smit2020chexbert,jain2021radgraph}.
\end{itemize}
\FloatBarrier

All metrics are reported as percentages.

\section{Additional Method and Implementation Details}
\label{app:method_details}

\subsection{Norm-clipped Dynamic Injection}
\label{app:norm_clipping}

In the main text, we describe DIVE as adding a state-conditioned residual to the hidden state. 
In implementation, we use norm-clipped addition to prevent large hidden-state magnitude shifts:
\[
    \tilde{h}_{t}^{(l,b)}
    =
    \left(h_{t}^{(l,b)}+\delta_{t}^{(l,b)}\right)
    \cdot
    \min\left(
    1,
    \frac{\rho \left\|h_{t}^{(l,b)}\right\|_2}
    {\left\|h_{t}^{(l,b)}+\delta_{t}^{(l,b)}\right\|_2}
    \right).
\]
Here $\rho>1$ controls the maximum allowed norm growth after injection.

\subsection{Teacher Construction and Logit Caching}
\label{app:teacher_cache}

For each training example, the teacher receives $N$ text-only report demonstrations, the query image, and the report-generation instruction. 
The teacher is evaluated under teacher forcing, conditioned on the target prefix $y_{<t}$ at each answer position. 
For efficiency, we cache the top-$K$ teacher logits at every supervised answer token, including the EOS position. 
During student training, the query-only student matches the cached teacher distribution over the teacher's top-$K$ support.

\subsection{Top-$K$ KL Renormalization}
\label{app:topk_kl}

Let $\mathcal{V}_t^K$ denote the teacher's top-$K$ token set at position $t$. 
We restrict both teacher and student distributions to $\mathcal{V}_t^K$ and renormalize them before computing KL divergence:
\[
p^T_{t,K}(v)=
\frac{p^T_t(v)}
{\sum_{u\in \mathcal{V}_t^K}p^T_t(u)},
\qquad
p^S_{t,K}(v)=
\frac{p^S_t(v)}
{\sum_{u\in \mathcal{V}_t^K}p^S_t(u)}.
\]
The KL loss is then computed over this restricted support.

\subsection{Inference and Decoding Configuration}
\label{app:decoding}

At inference time, the demonstration prompt and teacher cache are removed. 
The model receives only the query image and instruction prompt, with DIVE adapters activated at the selected decoder layers and branches. 
Unless otherwise specified, DIVE and all baselines use the same decoding configuration, including the same maximum number of new tokens and no post-hoc truncation.

\section{Clinical Finding Lexicon}
\label{app:finding_lexicon}

\begin{longtable}{p{0.22\textwidth}p{0.72\textwidth}}
\caption{Clinical finding phrase lexicon used for finding-weighted CE. For each sample, only labels selected by the CheXpert label policy are activated. Each phrase is tokenized with case variants and optional leading-space variants.}
\label{tab:finding_phrase_lexicon} \\
\toprule
\textbf{Finding label} & \textbf{Matched phrases} \\
\midrule
\endfirsthead

\toprule
\textbf{Finding label} & \textbf{Matched phrases} \\
\midrule
\endhead

Atelectasis &
atelectasis; atelectatic; volume loss; low lung volumes; low volume; subsegmental opacity; dependent atelectasis; plate-like atelectasis; platelike atelectasis \\

Cardiomegaly &
cardiomegaly; cardiac enlargement; enlarged heart; mild cardiomegaly; mildly enlarged heart; prominent cardiac silhouette; prominent cardiomediastinal silhouette \\

Consolidation &
consolidation; airspace disease; focal airspace disease; retrocardiac consolidation; left basilar consolidation; right basilar consolidation; bibasilar consolidation \\

Edema &
edema; pulmonary edema; interstitial edema; interstitial pulmonary edema; vascular congestion; pulmonary vascular congestion; pulmonary venous congestion; mild pulmonary edema; congestive heart failure; chf \\

Enlarged Cardiomediastinum &
enlarged cardiomediastinum; enlarged cardiomediastinal silhouette; mediastinal widening; prominent mediastinum; enlarged mediastinum; prominent cardiomediastinal silhouette \\

Fracture &
fracture; rib fracture; clavicle fracture; compression fracture; compression deformity; rib deformity \\

Lung Lesion &
lung lesion; pulmonary nodule; lung nodule; pulmonary mass; lung mass; nodular opacity; pulmonary nodular opacity; spiculated nodule \\

Lung Opacity &
opacity; opacities; lung opacity; hazy opacity; hazy opacities; bibasilar opacities; basilar opacities; retrocardiac opacity; left basilar opacity; right basilar opacity; airspace opacity; airspace opacities; interstitial opacities; patchy airspace opacity; patchy airspace opacities \\

No Finding &
no finding; no acute cardiopulmonary abnormality; no acute cardiopulmonary disease; no acute disease; no acute findings; clear lungs \\

Pleural Effusion &
pleural effusion; effusion; small effusion; small pleural effusion; trace effusion; trace pleural effusion; bilateral pleural effusions; costophrenic angle blunting; blunting of the costophrenic angle \\

Pleural Other &
pleural other; pleural thickening; pleural scarring; pleural plaque; pleural plaques \\

Pneumonia &
pneumonia; infectious infiltrate; infection; infectious process; pneumonic infiltrate \\

Pneumothorax &
pneumothorax; tiny pneumothorax; trace pneumothorax; small pneumothorax; apical pleural line \\

Support Devices &
support device; support devices; endotracheal tube; enteric tube; feeding tube; central venous catheter; central line; line tip; catheter tip; right ij; left ij; right internal jugular; left internal jugular; swan-ganz catheter; swan ganz catheter; port-a-cath; porta cath; right chest port; left chest port; tube tip; ett; et tube \\

\bottomrule
\end{longtable}

Table~\ref{tab:finding_phrase_lexicon} lists the phrase lexicon used for finding-weighted CE. Each CheXpert-style finding label is associated with a set of surface phrases. During training, we tokenize each phrase with case variants and optional leading-space variants, then upweight ground-truth answer tokens that match phrases from the activated finding categories.

\section{Ablation Study}

\subsection{Objective Components}
We further ablate the objective design on the LLaVA-Med backbone to separate the effects of KL distillation, decisive CE, and dynamic adaptation. As shown in Table~\ref{tab:llava-objective-ablation}, using KL with uniform CE preserves surface-level report quality but gives substantially weaker clinical correctness, indicating that ordinary teacher-forced CE under-allocates supervision to clinically decisive tokens. Conversely, removing KL and training only with decisive CE improves clinical signal over uniform CE, but degrades text overlap and graph-based factuality, suggesting that KL remains important for preserving the teacher's report distribution and generation structure.

The static-adapter variant also performs worse than full DIVE, even under the same DIVE objective. This indicates that the gain is not merely from token reweighting, but from combining clinically targeted supervision with an input-conditioned dynamic adapter. Overall, full DIVE gives the strongest balance across lexical, semantic, and clinical metrics, supporting the need for both dynamic adaptation and the combined KL plus decisive-CE objective.

\begin{table}[t]
\centering
\caption{Objective ablation on LLaVA-Med. All metrics are reported on the CheXpert Plus test set.}
\label{tab:llava-objective-ablation}
\resizebox{\linewidth}{!}{
\begin{tabular}{lccccccc}
\toprule
Method & B-1 & B-4 & R-L & CIDEr & BERTScore & CheXbert & RadGraph \\
\midrule
Dynamic + KL + uniform CE & 23.92 & 4.74 & 18.11 & 6.07 & 85.70 & 30.67 & 14.68 \\
Dynamic + decisive CE, no KL & 14.92 & 2.89 & 16.51 & 5.05 & 85.61 & 35.38 & 14.58 \\
Static + DIVE objective & 18.65 & 3.76 & 17.82 & 5.52 & \textbf{85.99} & 26.77 & 13.80 \\
DIVE (Ours) & \textbf{27.79} & \textbf{5.31} & \textbf{18.99} & \textbf{7.49} & 85.86 & \textbf{43.57} & \textbf{16.88} \\
\bottomrule
\end{tabular}
}
\end{table}

\subsection{Weight Sensitivity}
\label{app:weight_sensitivity}

We further examine the sensitivity of DIVE to the two decisive-token weights:
the clinical finding CE weight $\omega_{\mathrm{path}}$ and the EOS weight
$\omega_{\mathrm{EOS}}$. 
For the clinical-weight sweep, we fix $\omega_{\mathrm{EOS}}=5$ and vary
$\omega_{\mathrm{path}}$. 
For the EOS-weight sweep, we fix $\omega_{\mathrm{path}}=8$ and vary
$\omega_{\mathrm{EOS}}$.

Table~\ref{tab:llava-weight-sweep} reports the results on CheXpert Plus with
the LLaVA-Med backbone. 
For clinical finding supervision, very small weights provide limited clinical
benefit, while moderate weights improve both lexical and clinical-proxy metrics.
The best CheXbert score is obtained at $\omega_{\mathrm{path}}=8$, although
nearby settings such as $\omega_{\mathrm{path}}=5$ give comparable BLEU-4 and
RadGraph performance. 
Increasing the weight to $\omega_{\mathrm{path}}=10$ degrades both lexical and
clinical-proxy metrics, suggesting that excessive emphasis on finding tokens can
distort generation.

EOS weighting shows a similar trade-off. 
Small EOS weights underperform on clinical-proxy metrics, whereas
$\omega_{\mathrm{EOS}}=5$ and $\omega_{\mathrm{EOS}}=8$ form a relatively stable
high-performing range. 
The default setting, $\omega_{\mathrm{path}}=8$ and $\omega_{\mathrm{EOS}}=5$,
achieves the best CheXbert score and competitive RadGraph performance, indicating
a balanced choice rather than a highly sensitive optimum.

\begin{table}[t]
\centering
\small
\caption{
Weight sensitivity of DIVE with LLaVA-Med on CheXpert Plus.
The clinical-weight sweep fixes $\omega_{\mathrm{EOS}}=5$, and the EOS-weight
sweep fixes $\omega_{\mathrm{path}}=8$.
}
\label{tab:llava-weight-sweep}
\resizebox{\linewidth}{!}{
\begin{tabular}{lccccccc}
\toprule
Weight setting & B-1 & B-4 & R-L & CIDEr & BERTScore & CheXbert & RadGraph \\
\midrule
\multicolumn{8}{l}{\textit{Clinical finding CE weight sweep ($\omega_{\mathrm{EOS}}=5$)}} \\
$\omega_{\mathrm{path}}=0$  & 14.84 & 3.11 & 17.55 & 4.89 & 86.16 & 28.12 & 14.59 \\
$\omega_{\mathrm{path}}=1$  & 14.82 & 3.10 & 17.55 & 4.90 & 86.16 & 27.75 & 14.58 \\
$\omega_{\mathrm{path}}=3$  & 24.98 & 5.11 & \textbf{19.27} & \textbf{8.13} & \textbf{86.17} & 40.46 & 16.91 \\
$\omega_{\mathrm{path}}=5$  & 27.26 & \textbf{5.35} & 19.21 & 8.01 & 86.03 & 42.50 & \textbf{17.02} \\
$\omega_{\mathrm{path}}=8$  & \textbf{27.79} & 5.31 & 18.99 & 7.49 & 85.86 & \textbf{43.57} & 16.88 \\
$\omega_{\mathrm{path}}=10$ & 23.28 & 4.38 & 18.21 & 6.41 & 85.17 & 38.97 & 15.69 \\
\midrule
\multicolumn{8}{l}{\textit{EOS weight sweep ($\omega_{\mathrm{path}}=8$)}} \\
$\omega_{\mathrm{EOS}}=0$ & 22.76 & 4.69 & 18.75 & 7.62 & 85.98 & 35.49 & 15.42 \\
$\omega_{\mathrm{EOS}}=1$ & 21.77 & 4.68 & 18.73 & 7.63 & 85.99 & 37.77 & 15.50 \\
$\omega_{\mathrm{EOS}}=2$ & 23.56 & 4.77 & 19.01 & 7.85 & \textbf{86.04} & 36.52 & 15.59 \\
$\omega_{\mathrm{EOS}}=3$ & 26.61 & 4.99 & 18.74 & 7.25 & 85.67 & 41.86 & 16.47 \\
$\omega_{\mathrm{EOS}}=5$ & \textbf{27.79} & \textbf{5.31} & 18.99 & 7.49 & 85.86 & \textbf{43.57} & \textbf{16.88} \\
$\omega_{\mathrm{EOS}}=8$ & 27.01 & 5.15 & \textbf{19.02} & \textbf{7.88} & 85.89 & 43.21 & 16.28 \\
\bottomrule
\end{tabular}
}
\end{table}

\section{Qualitative Examples}
\label{app:qualitative_examples}

\begin{figure*}[t]
    \centering
    \includegraphics[width=0.98\textwidth]{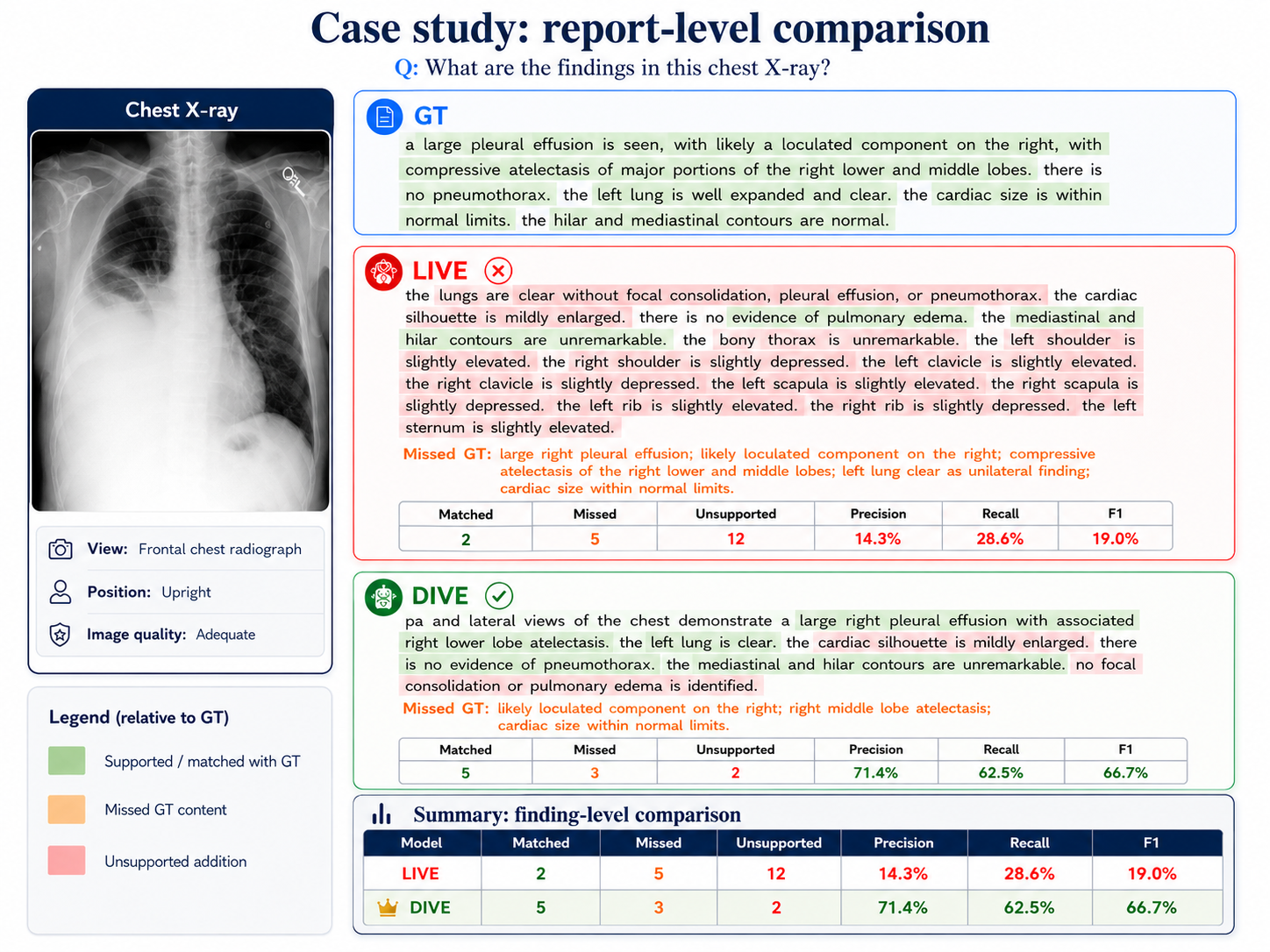}
    \caption{
    \textbf{Qualitative case study on a large right pleural effusion.}
    The GT report describes a large right pleural effusion with a likely loculated component and compressive atelectasis of the right lower and middle lobes.
    LIVE misses the main abnormality and generates many unsupported anatomical-position statements.
    DIVE captures the large right pleural effusion and associated right lower lobe atelectasis, while still missing finer-grained details such as the loculated component and right middle lobe involvement.
    Green highlights denote findings supported by the GT report, orange highlights denote missed GT content, and red highlights denote unsupported additions.
    The summary table reports finding-level matched, missed, and unsupported counts, together with precision, recall, and F1.
    }
    \label{fig:qual_case_effusion}
\end{figure*}

Figure~\ref{fig:qual_case_effusion} provides a qualitative example comparing LIVE and DIVE on a chest X-ray with a large right pleural effusion.
The GT report describes a large pleural effusion with a likely loculated component on the right, compressive atelectasis involving major portions of the right lower and middle lobes, no pneumothorax, a clear left lung, normal cardiac size, and normal hilar and mediastinal contours.
We annotate each generated report at the phrase level: green highlights denote findings supported by the GT report, orange highlights denote missed GT content, and red highlights denote unsupported additions.

LIVE fails to recover the main abnormality in the image.
Instead of identifying the large right pleural effusion and associated compressive atelectasis, it generates a largely normal report and introduces many unsupported anatomical-position statements, including statements about the shoulders, clavicles, scapulae, ribs, and sternum.
This leads to a large number of unsupported additions and poor finding-level precision and F1.

DIVE better captures the clinically salient abnormality by identifying a large right pleural effusion with associated right lower lobe atelectasis.
It also preserves several key negative or normal findings, including no pneumothorax, a clear left lung, and unremarkable mediastinal and hilar contours.
However, DIVE still misses finer-grained GT details, including the likely loculated component, right middle lobe involvement, and normal cardiac size.
It also introduces unsupported statements such as mild cardiac enlargement and absence of focal consolidation or pulmonary edema.
Overall, this example shows that DIVE improves alignment with the major abnormal finding and reduces unsupported over-generation compared with LIVE, while still leaving room for better fine-grained clinical grounding.

%%%%%%%%%%%%%%%%%%%%%%%%%%%%%%%%%%%%%%%%%%%%%%%%%%%%%%%%%%%%

%%%%%%%%%%%%%%%%%%%%%%%%%%%%%%%%%%%%%%%%%%%%%%%%%%%%%%%%%%%%

\clearpage

\end{document}